\pdfoutput=1
\documentclass[11pt]{article}
\usepackage[numbers]{natbib}

\usepackage[final]{acl}

\usepackage{times}
\usepackage{latexsym}
\usepackage{soul} % for highlighiting text

\usepackage{amsmath}    % For advanced math symbols and environments
\usepackage{amssymb}    % For additional math symbols like \mathbb
\usepackage{amsfonts}   % For fonts like \mathbb (e.g., for the field of scalars)
\usepackage{tikz-cd}    % For drawing commutative diagrams and string diagrams
\usepackage{stmaryrd}   % For additional symbols if needed (like maps and arrows)

\usepackage{hyperref}

\usepackage[T1]{fontenc}
\usepackage[utf8]{inputenc}
\usepackage{microtype}
\usepackage{inconsolata}

\usepackage{graphicx}
\usepackage{amsmath}   % For mathematical expressions
\usepackage{braket}    % For Dirac notation
\usepackage{subcaption} % For subfigures
\usepackage{colortbl}  % For coloring cells
\usepackage{booktabs}  % For better table lines
\usepackage{multirow}
\usepackage{multicol}

\definecolor{lightgray2}{rgb}{0.9, 0.9, 0.9}

\title{Multimodal Structure-Aware Quantum Data Processing}

\author{Hala Hawashin \\
University College London\\
\texttt{hala.hawashin.23@ucl.ac.uk} \\\And
Mehrnoosh Sadrzadeh \\
University College London \\
\texttt{\textcolor{black}{m.sadrzadeh@ucl.ac.uk}}\\}

\begin{document}
\maketitle

\begin{abstract}
%Hala's Original: The current state of  large language models have advanced the field of natural language processing, but their "black box" nature limits interpretability. To address this, researchers proposed Quantum Natural Language Processing (QNLP), which provides a translation between language constructions and quantum circuits and has been experimented with on language tasks. To this end, we extend QNLP by (a) enriching its theoretical framework with types and type homomorphisms that model images and image properties (b) developing compositional architectures that combine images and text in different ways. In this paper, we develop \textit{MultiQ}-NLP, short for Multimodal Quantum NLP: a framework for structure-aware data processing with language and other sources of data such as images. We test our framework on a main stream image classification task released by Google. Our comparative analysis of four compositional models reveals that incorporating the structure of language enhances learning transparency and improves prediction accuracy. While our quantum model performs on par with state-of-the-art classical models, it demonstrates that structured models outperform bag-of-words approaches.
%What follows is  the introduction of the QIP extended abstract
While large language models (LLMs) have advanced the field of natural language  processing (NLP), their ``black box'' nature obscures their decision-making processes. To address this, researchers developed structured approaches using  higher order tensors. These are able to model linguistic relations, but stall when training  on classical computers due to their excessive size. Tensors are natural inhabitants of quantum systems and training  on quantum computers provides a solution by translating text to  variational quantum circuits. In this paper, we develop \emph{MultiQ}-NLP: a framework for structure-aware data processing  with multimodal text+image data. \textcolor{black}{Here, ``structure'' refers to syntactic and grammatical relationships in language, as well as the hierarchical organization of visual elements in images.} We enrich the translation with new types and type homomorphisms  and develop novel architectures to represent structure. When tested on a main stream image classification task (SVO Probes), our best  model showed a par performance with the state of the art classical models; moreover the best model was fully structured\footnote{This paper originated from the MSc thesis of the first author, which was awarded distinction by UCL Computer Science in September 2024 \cite{hala_thesis}. The thesis and its associated code can be downloaded from \href{http://arxiv.org/abs/2411.05023}{here}.}. 
\end{abstract}

\section{Introduction}
Quantum computing is an emerging technology capable of solving complex problems beyond the reach of classical computers. By combining principles of quantum mechanics with computer science, it significantly impacts various fields, including Natural Language Processing (NLP). Recent advances in NLP, particularly through state-of-the-art transformer models, have led to the rise of large language models (LLMs) which are transforming our day to day lives. However, a persistent challenge in this is the ``black box'' dilemma, where the complex nature of deep neural networks that tune millions of parameters result in opaque decision-making processes that are hard to be interpreted.

% Dimitri notes:  where the opaque decision-making processes of these models result from the complex deep neural networks that tune millions of parameters. }

To address this, researchers developed structured approaches that rely on higher-order tensors \cite{baroni-zamparelli-2010-nouns,mathfoundation2010,grefenstette-sadrzadeh-2011-experimental,kartsaklis-etal-2012-unified,maillard-etal-2014-type,wijnholds-etal-2020-representation}. These approaches can model linguistic relations via tensor contraction but face limitations when it comes to training on classical computers due to the computational complexity. Since tensors are well-suited to quantum systems, training on quantum computers has been proposed as a solution \cite{QNLPinPractice,Meichanetzidis:2020qcw,wazni2024large,wazni-sadrzadeh-2023-towards,widdows2024near}. This idea leverages the compositional structure shared by both languages and quantum systems, which can be modelled within the same mathematical framework. This has led to the development of \texttt{lambeq}, a high-level programming toolkit for Quantum Natural Language Processing (QNLP) \cite{lambeq2021}. Despite its potential, QNLP has primarily focused on textual data, ignoring other modalities such as images, videos and audio.

% Dimitri notes: falling behind classical NLP which integrates various modalities, including images, video, and audio, to create Multimodal NLP.

In this paper, we introduce \emph{MultiQ}-NLP, short for Multimodal Quantum NLP, the first framework for structure-aware data processing that integrates language and other data sources, such as images. To this end, we extend existing QNLP models by (a) enriching its theoretical framework with types and type homomorphisms to model images and image properties, and (b) developing compositional architectures that combine images and text in innovative ways. Our comparative analysis of four different compositional models reveals that incorporating the compositional structure of language enhances training transparency and improves prediction accuracy. We test our framework on mainstream image classification tasks (see Section \ref{sec: task}) to evaluate its sensitivity to verb semantics and robustness in handling syntactic variations like interchangeable subjects and objects.

% Dimitri notes: a mainstream image classification task \cite{Google-SVO}. 

The central hypothesis of this study is that syntactic, structure-aware models will outperform those lacking syntactic information by leveraging linguistic patterns for more accurate interpretations. This is tested in two ways: first, by assessing the models’ ability to capture distinctions in verb usage and accurately interpret actions; and second, by evaluating how well the models handle syntactic variations, such as differentiating between interchangeable subjects and objects within sentence structures.

% The central hypothesis of this study is that syntactic, structure-aware models will outperform those without syntactic information by leveraging linguistic patterns for more accurate interpretations. This is tested in two ways: first, using a sub-dataset curated from \cite{Google-SVO} to focus on verb usage; and second, with a manually curated dataset designed to evaluate the handling of interchangeable subjects and objects within sentence structures.

%  Dimitri notes: Additionally, the study investigates a sub-hypothesis concerning verb composition and understanding. This aspect examines whether the models' ability to differentiate between sentence structures and verb usage significantly enhances their accuracy in interpretation. 

\section{\textcolor{black}{Related Work}}

A pivotal work in this area is \cite{mathfoundation2010} which introduces a method for employing quantum circuits to model the compositional structure of language. This approach effectively captures how individual word meanings combine to create sentence-level interpretations, aligning with categorical models in NLP that view grammar as a compositional structure. With its theoretical basis deeply embedded in category theory, this framework offers a more interpretable and structured approach to language processing. Monoidal categories, a key concept within category theory, are essential for representing linguistic compositionality in this context. 

The approach is further grounded in pregroup grammar, an algebraic structure introduced by \cite{lambek1999}, which is fundamental for modelling syntactic structures. \cite{bobcoecke2010categorical} highlighted a significant challenge in NLP: while dictionaries exist for individual words, there is a lack of similar resources for entire sentences. This gap is effectively addressed by Lambek’s work which assigns types to words and uses reductions to verify grammatical correctness, forming the backbone of syntactic analysis in the Distributional Compositional Categorical (DisCoCat) model.

The earliest application of DisCoCat on a Noisy Intermediate-Scale Quantum (NISQ) processor was demonstrated in the paper “Grammar-aware Sentence Classification on Quantum Computers” \cite{grammareAware}, using IBM's quantum hardware. This study encoded sentences as parametrized quantum circuits and used entangling operations to model grammatical structures. The results demonstrated that quantum circuits could handle NLP tasks with expressiveness and precision comparable to, and sometimes surpassing, classical methods. Very similar further work  followed suit, see \cite{QNLPinPractice,Meichanetzidis:2020qcw,wazni2024large,wazni-sadrzadeh-2023-towards,widdows2024near}.  The open source software package  \texttt{Lambeq} \cite{lambeq2021}, was introduced  to use in quantum machine learning. This toolkit inputs text from the user, translates it into variational quantum circuits, allows the user to input training and test data, then offers a variety of learning algorithms to learn the parameters. Lambeq works with the grammatical typed structures underlying  the text, encodes those in the parameters and structure of quantum circuits, thus captures semantic meaning more effectively than other quantum approaches to  language processing such as  \cite{QML,basile-tamburini-2017-towards,blacoe-etal-2013-quantum}.

\section{\textcolor{black}{Task}}
\label{sec: task}

The  task we work with in this study is a  classification task designed to assess the performance of a set of compositional structure-aware models on multimodal data. The general goal of this task is to  match images with their correct captions. We work with  a subset of Google's SVO-Probes dataset \cite{Google-SVO}, developed to demonstrate the abilities of   compositional approaches to image classification \cite{wazniverbclip}. We use contrastive learning where one learns from both positive and negative examples. SVO stands for Subject-Verb-Object.  The verbs of an SVO sentence are the centre point of composition, they have a tensorial representation that allows them to relate the meaning of the subject to that of the object.  Since the focus our work is on composition, in one if our tasks we restrict the contrastive learning to those negative examples where the verbs differ.  In our second task, we focus on the general structure of the captions, keep the verb the same and swap the subject and subject. The  details of this task,  their associated datasets, and examples are provided in  Section 5 on Data Collection below.

\section{Methodology}

This section explains the diagrammatic representation and transformation pipeline for how text and image data is represented diagrammatically and subsequently transformed into quantum circuits.

\subsection{Modelling Linguistic Structure in QNLP}
\label{Modelling Linguistic Structure in QNLP}

 At its core, the mathematical model behind QNLP is theory of categories, consisting of three fundamental components: objects, morphisms, and composition. 
 \begin{itemize}
     \item \textit{Objects} \( A, B, C, \dots \) represent types or entities and serve as foundational elements. In QNLP, they represent grammatical and semantic types.
    
     \item \textit{Morphisms} \( f: A \to B \) denote processes that map one object to another, illustrating how entities transform. In QNLP, these represent grammatical and semantic reductions.
    
     \item \textit{Composition} involves chaining morphisms together. For instance, given morphisms \( f: A \rightarrow B \) and \( g: B \rightarrow C \), their composition \( g \circ f: A \rightarrow C \) is represented by arrows from \( A \) to \( B \) and from \( B \) to \( C \). In QNLP, it models the linguistic rules of syntax and semantics.
 \end{itemize}

At its core, the mathematical model behind compositional models is theory of categories. Compact categories  have a tensor product  $A \otimes B$ with a left and a right adjoint, denoted by $l$ and $r$ subscripts. As an example, consider a category with objects  the basic grammatical types of noun and sentence: $n$ and $s$. Tensor forms  complex types such as those for adjectives $n^r \otimes  s$ and verbs $n^r \otimes s \otimes n^l$. Given the map-state duality, each tensor is also a map and tensor contractions stand for function-argument applications.

% Mehrnoosh Notes: \hl{GIVEN THE MAP-STATE DUALITY, EACH SUCH TENSOR IS ALSO A MAP AND TENSOR CONTRACTION STANDS FOR FUNCTION-ARGUMENT APPLICATION)}. 

An example of composition is the construction of a transitive sentence, e.g.   "Dogs chase cats"; this sentence is modelled by the tensor  $ n \otimes  (n^r \otimes s \otimes n^l) \otimes n$, which reduces to the atomic type $s$. 

\begin{figure}[h!]
    \centering
    \includegraphics[width=0.8\linewidth]{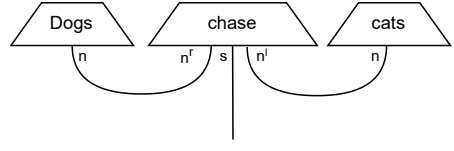}
    \caption{The \texttt{QNLP} string diagram representing the sentence "Dogs chase cats".}
    \label{fig:DisCoCat-diag}
\end{figure}

Compact categories have a complete string diagrammatic calculus depicting their computations. Therein, objects are boxes and morphisms are lines between the boxes. Composition connects lines when their types match, while tensoring corresponds to juxtaposition. Special morphisms, called \textit{epsilon maps}, encode function-argument application and are represented diagrammatically by cups.  \textcolor{black}{The epsilon map, defined as:
\[\epsilon: V \otimes V^* \rightarrow \mathbb{I}\]
where \( V \) is a vector space, \( V^* \) is its dual, and \( \mathbb{I} \) is the scalar field, collapses to produce a scalar. When applied, this contraction helps connect function-argument representations.} As an example, we provide the diagrammatic representation of a transitive sentence in Figure \ref{fig:DisCoCat-diag}. \textcolor{black}{Here, the cups depict the application of the map corresponding to the tensor of the transitive verb ("Chase") to the vectors of the subject ("Dogs") and object ("Cats").} 

%See Figure \ref{fig:Sim14-Diag-wires} for an overview of the diagrammatic calculus.

\textcolor{black}{In the transitive sentence "Dogs chase cats",  \( \epsilon \) contracts each noun vector ("Dogs" and "Cats") with the verb tensor ("Chase") through function-argument application, while \( \text{id}_N \) serves as an identity map that preserves the structure of the nouns as they connect to the verb. This morphism sequence thus forms a single vector in the sentence space \( S \) that encodes the entire sentence meaning.\[\epsilon \circ (\text{id}_N \otimes \epsilon) \circ (\text{id}_N \otimes \text{Chase} \otimes \text{id}_N)\]}

% Mehrnoosh Notes: \hl{SPECIAL MORPHISMS CALLED EPSILON MAPS ENCODE FUNCTION-ARGUMENT APPLICATION. THESE ARE DENOTED BY CUPS DIAGRAMMTICALLY. SEE FIGURE ??? FOR THE BASICS OF THE DIAGRAMMATIC CALCULUS. AS AN EXAMPLE, WE PROVIDE THE} diagrammatic representation OF a transitive sentence \hl{IN Figure \ref{fig:DisCoCat-diag}. AS YOU CAN SEE, THE CUPS DEPICT THE APPLICATION OF THE MAP CORRESPONDING TO THE TENSOR OF THE TRANSITIVE VERB (CHASE) TO THE VECTORS OF THE SUBJECT (DOGS) AND  OBJECT (CATS).} TODO: Check if the figure referencing correct?

\begin{figure*}[h!]
    \centering
    \includegraphics[width=\textwidth]{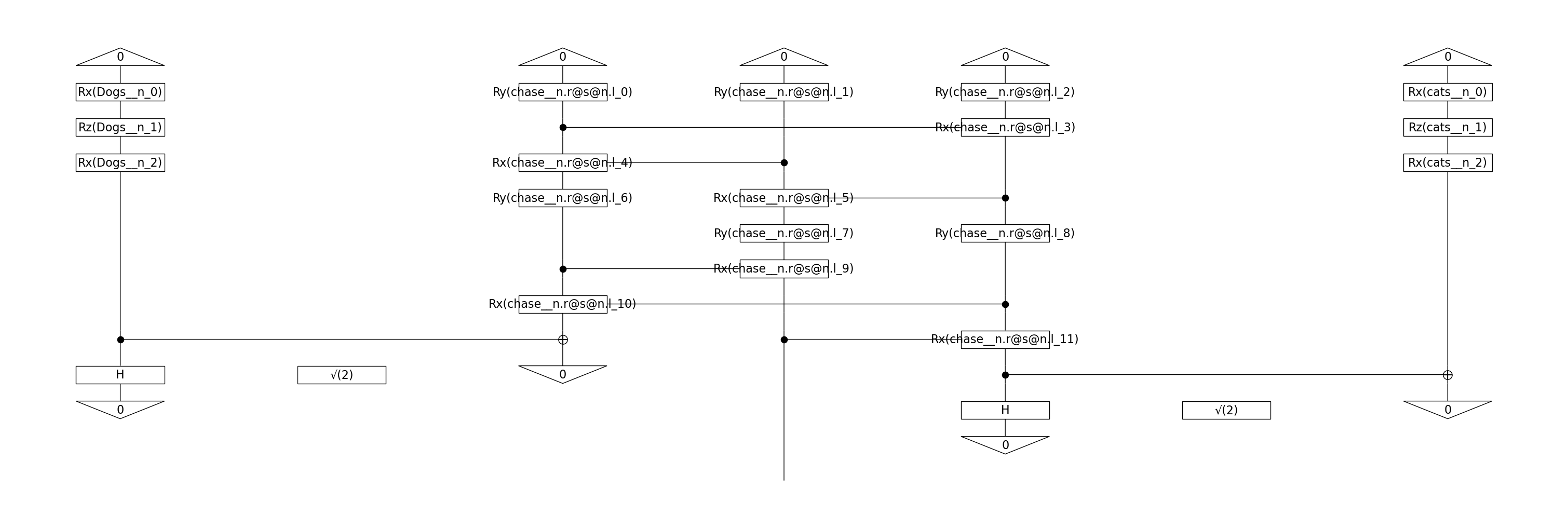}
    \caption{The quantum circuit representation for the sentence "Dogs chase cats".}
    \label{fig:DisCoCat-circ}
\end{figure*}

% \begin{figure}[h!]
%     \centering
%     \includegraphics[width=1.0\linewidth]{Diagram2/sentence-circ.png}
%     \caption{The quantum circuit representation for the sentence "Dogs chases cats".}
%     \label{fig:DisCoCat-circ}
% \end{figure}

In many ways, there is a  parallel between the compositional nature of language, where words interact and combine to form sentences, to the structure of quantum circuits, where quantum gates act on qubits and yield results for computations. Thus, the compositional aspects of language can be modelled like quantum computation, with operations (morphisms) applied to quantum states (objects) in a structured manner. This idea forms the basis of compositional NLP's text-to-circuit translation, where different ans\"atze are applied to assign a parametrised quantum circuit to  grammatical types and linguistic reductions. \textcolor{black}{Here, we use the \texttt{Sim14} ansatz, where (i) rotation gates are used to model qubits, representing individual words by capturing different aspects of their meaning; (ii) controlled rotation gates form tensor states, reflecting the interaction between words, such as how verbs influence nouns; (iii) combinations of CNOT and post-selection are used to model grammatical reductions, simplifying phrases while preserving semantic coherence; and (iv) CNOT is used as the logical operation for conjunction, maintaining the connection between words or phrases in a sentence. The intuition behind these operations is to mirror the compositional structure of language within a quantum framework. Just as words combine and reduce meaningfully in sentences, these quantum operations enable structured transformations of quantum states, preserving both syntax and semantics in the process.}

% ^ Mehrnoosh Notes: Here, we use the \texttt{Sim14} ansatz, where (i)  rotation gates  are used to model  qubits, (ii) controlled rotation gates are used to  form  tensor states, (iii) combinations of CNOT and post-selection are used to model grammatical reductions, and (iv) CNOT is used as the logical operations of conjunction. \hl{HALA DO YOU THINK YOU CAN EXPLAIN WHY THESE OPERATIONS ARE USED? WHAT IS THE INTUITION BEHIND THEM?} 

\begin{figure}[h!]
    \centering
    \includegraphics[width=1.0\linewidth]{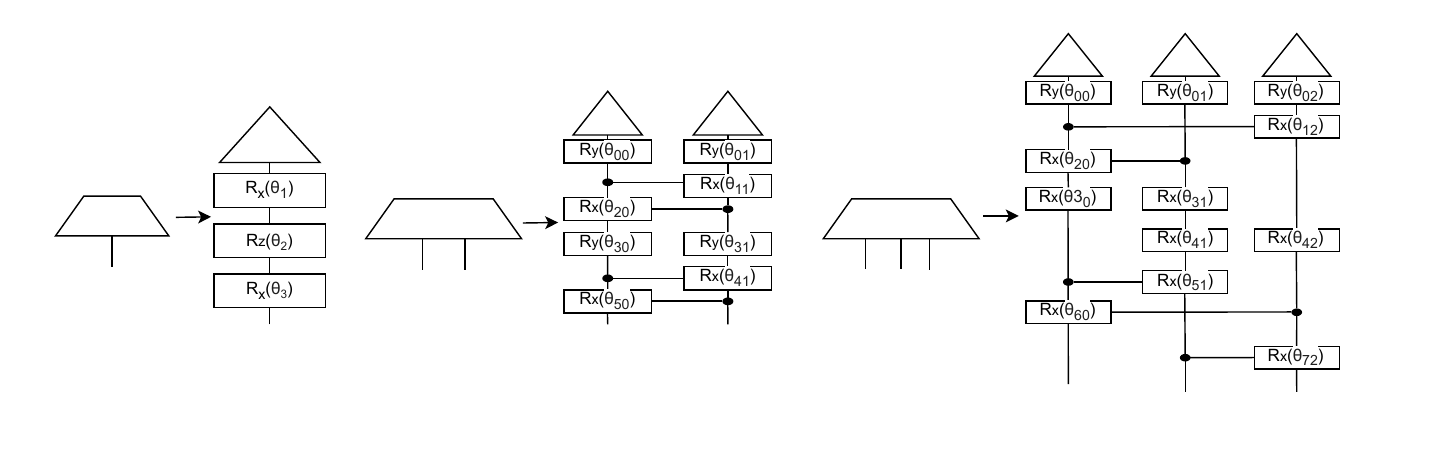}
    \caption{String diagram translation using Sim14 ansatz.}
    \label{fig:Sim14-Diag}
\end{figure}

\begin{figure}[h!]
    \centering
    \includegraphics[width=1.0\linewidth]{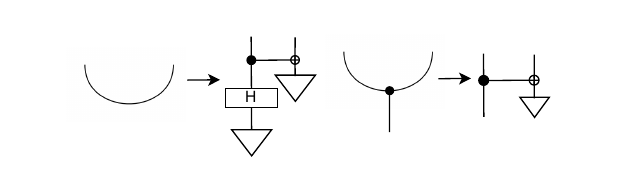}
    \caption{Wires in string diagram translation using Sim14 ansatz.}
    \label{fig:Sim14-Diag-wires}
\end{figure}

\textcolor{black}{This translation is depicted in Figures \ref{fig:Sim14-Diag} and \ref{fig:Sim14-Diag-wires}, with an example circuit for the sentence "Dogs chase cats" shown in Figure \ref{fig:DisCoCat-circ}. Figure \ref{fig:Sim14-Diag} illustrates how the string diagram is gradually decomposed into quantum operations. As we move from left to right, the diagram shows how rotations accumulate, representing the contribution of each word to the overall meaning of the sentence. These rotations are parametrised based on the linguistic model, aligning directly with the compositional structure of the language. In the quantum translation, the cup---originally representing the function-argument morphism in the string diagram---becomes a series of quantum gates that encode this interaction (see the left side of Figure \ref{fig:Sim14-Diag-wires}). A distinct morphism called the "spider" represents a Frobenius multiplication where its corresponding quantum translation are depicted in the right side of Figure \ref{fig:Sim14-Diag-wires}.}

% \begin{figure}[h!]
%     \centering
%     \includegraphics[width=1.0\linewidth]{Diagram2/Sim14-circuits.pdf}
%     \caption{String diagram translation using Sim14 ansatz.}
%     \label{fig:Sim14-Diag}
% \end{figure}

\subsection{Modelling Images and Language in MultiQ-NLP}

The categorical representation of natural language was developed in \cite{Lambek1958-LAMTMO-5,Lambek1988} and is the free compact closed category generated over a set of predefined atomic grammatical types. In order to reason about images, we expand our set of generators with two  new atomic types and their corresponding morphisms: one  for images called $img$ and one for  the combination of linguistic and image data  call a  \texttt{MultiQ\_box} (labelled as \texttt{COMPARISON} in the diagrams below).

\begin{figure}[h!]
    \centering
    \includegraphics[width=1.0\linewidth]{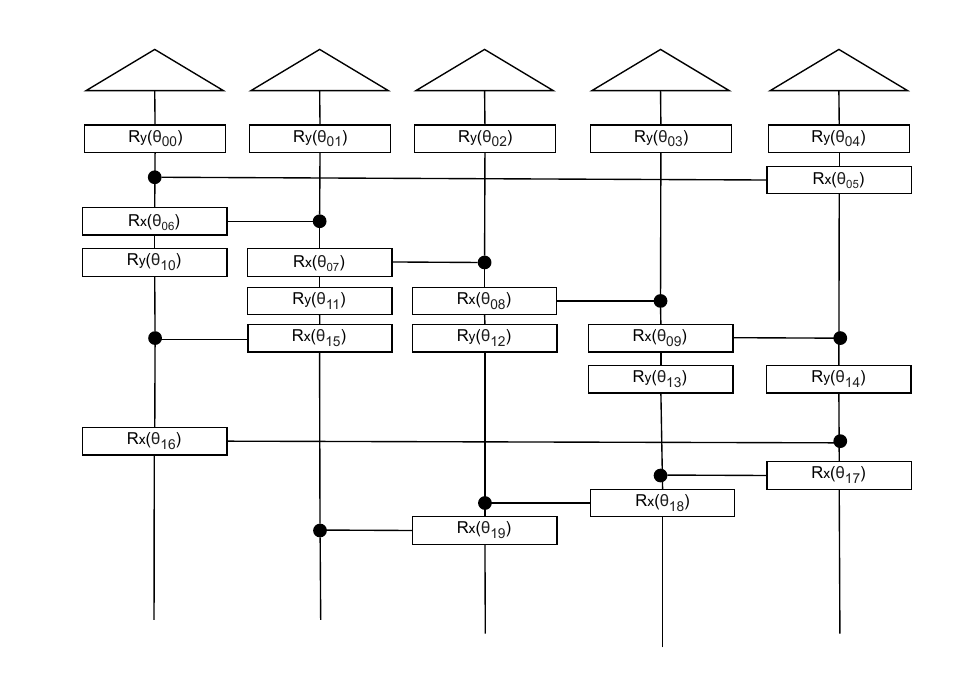}
    \caption{The quantum circuit representation for \texttt{img} using Sim14 ansatz.}
    \label{fig:Sim14-Image}
\end{figure}

The $img$ type is translated into a set of qubits which are sequentially initialised by the data encoded in the  feature vectors of the images, learnt by classical machine learning algorithms. Figure~\ref{fig:Sim14-Image} shows the quantum circuit translation of the image feature vector, where each parameterized rotation gate (e.g., $\theta_{00}$, $\theta_{01}$, etc.) corresponds to image-derived qubit states. The circuit assigns qubits based on the dimensionality of the input image features. Rotation gates adjust the qubits' states to store the image feature vectors, capturing variations in the quantum state space. Controlled operations introduce entanglement to maintain dependencies among features, while the layered structure enables sequential processing and compositional interactions. This approach transforms high-dimensional image data into a structured quantum form, facilitating multimodal integration.

% ^ Mehrnoohs Notes: ADD ALSO SOME EXPLANATIONS TO THE MAIN IMAGE IN FIGURE 6 AND THE IMAGE IN FIGURE 5.}

The \texttt{MultiQ\_box} box integrates the linguistic circuits with the image qubits and outputs a set of probabilities. These probabilities stand for the likelihoods that the images correctly depict a piece of text.  

% \newpage

\begin{figure}[h!]
    \centering
    \includegraphics[width=1.0\linewidth]{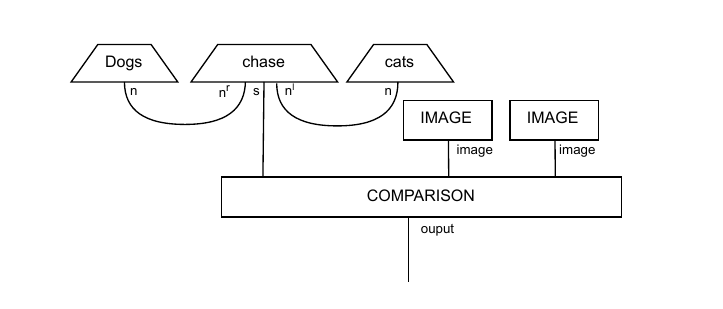}
    \caption{String diagrams for the sentence ``Dogs chase cats'' using \texttt{Cat}.}
    \label{fig:Cat-diag}
\end{figure}

\begin{figure}[h!]
    \centering
    \includegraphics[width=0.8\linewidth]{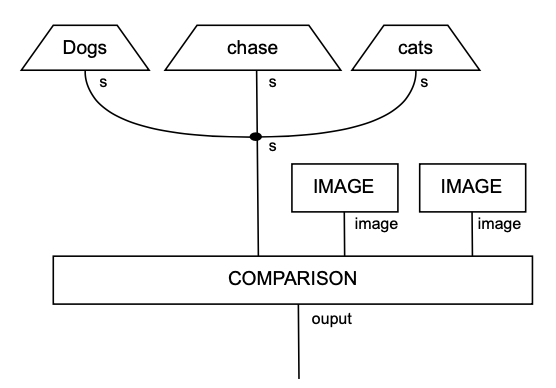}
    \caption{String diagrams for the sentence ``Dogs chase cats'' using \texttt{Bag-of-Words}.}
    \label{fig:BagOfWords-diag}
\end{figure}

\begin{figure}[h!]
    \centering
    \includegraphics[width=1\linewidth]{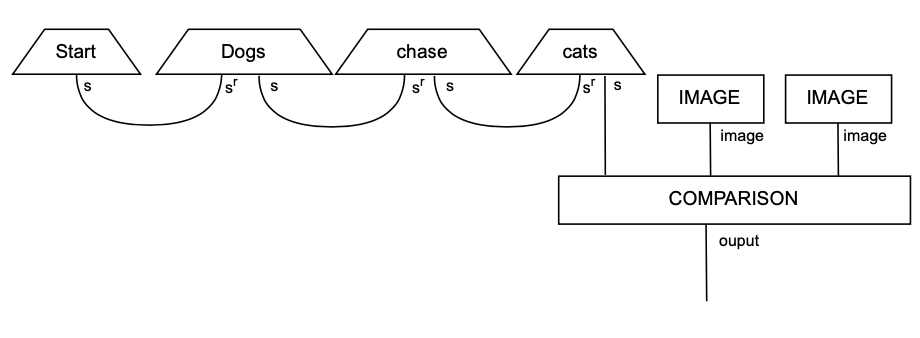}
    \caption{String diagrams for the sentence ``Dogs chase cats'' using \texttt{Word-Sequence}.}
    \label{fig:WordSequence-diag}
\end{figure}

\begin{figure}[h!]
    \centering
    \includegraphics[width=1\linewidth]{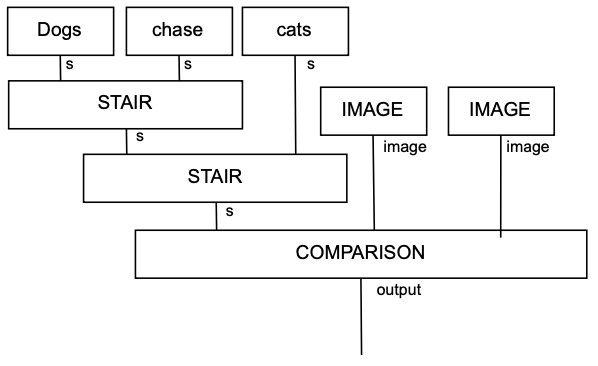}
    \caption{String diagrams for the sentence ``Dogs chase cats'' using \texttt{LTree}.}
    \label{fig:LTree-diag}
\end{figure}

\begin{figure}[h!]
    \centering
    \includegraphics[width=1.0\linewidth]{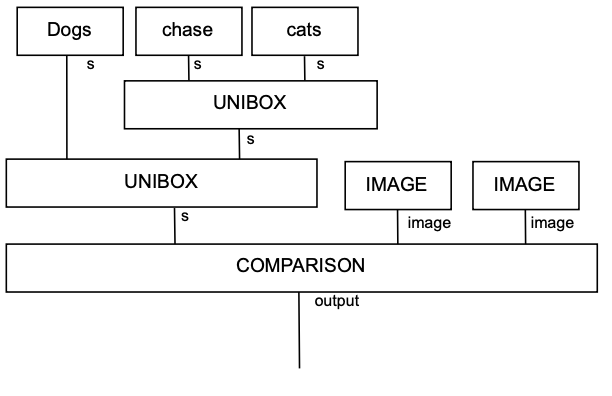}
    \caption{String diagrams for the sentence ``Dogs chase cats'' using \texttt{CFG}.}
    \label{fig:CFG-diag}
\end{figure}

Figures~\ref{fig:Cat-diag}, \ref{fig:BagOfWords-diag}, \ref{fig:WordSequence-diag}, \ref{fig:LTree-diag}, and \ref{fig:CFG-diag} illustrate the diagrammatic translation of combined text and image data, where image qubits and linguistic circuits merge through the \texttt{MultiQ\_box}. The integration of text-based grammatical reductions and image data into a single quantum circuit allows for a multimodal processing approach, ultimately outputting probabilities that represent the coherence between text and image inputs. 

The \texttt{Cat} model (see Figure \ref{fig:Cat-diag}) integrates distributional semantics with categorical compositionality to model the meaning of natural language. It uses functors to map diagrams from the category of pregroup grammars to vector space semantics. The process involves generating a Combinatory Categorial Grammar (CCG) derivation, which assigns grammatical categories to words and applies combinatory rules to combine these categories into larger syntactic units.

The \texttt{Bag-of-Words} model (see Figure \ref{fig:BagOfWords-diag}), in contrast, treats sentences as unordered collections of words, focusing solely on word occurrences and frequencies without considering syntax or grammatical relationships. 

The \texttt{Word-Sequence} model (see Figure \ref{fig:WordSequence-diag}) maintains the order of words in a sentence, representing sequential dependencies. Word vectors are fed into the quantum circuit in the same order as they appear in the sentence, allowing the model to capture the influence of word order on sentence meaning. Unlike the \texttt{Bag-of-Words} model, the Word Sequence model incorporates sequential structure but does not account for hierarchical grammatical relationships.

The \texttt{LTree} model (see Figure \ref{fig:LTree-diag}) introduces a more complex representation by incorporating a syntactic tree structure, where sentences are parsed into hierarchical trees that represent grammatical relationships. Each node in the tree corresponds to an operation that combines smaller units into larger syntactic constructs. This model allows for a more detailed syntactic analysis, capturing dependencies between components in a hierarchical manner.

The \texttt{CFG} model (see Figure \ref{fig:CFG-diag}) also leverages CCG derivations, similar to the \texttt{Cat} model. However, the \texttt{CFG} model captures hierarchical relationships directly, allowing for a straightforward interpretation of compositional steps. It eliminates the need for additional transformations, such as those required by pregroup grammar, making it more efficient in modeling recursive syntactic structures. Unlike the \texttt{Cat} model, which relies on pregroup grammar and morphisms for compositional analysis, the \texttt{CFG} model uses a tree-like representation that directly maps grammatical rules onto hierarchical structures.

% ^ Mehrnoohs Notes: \hl{HALA IT WOULD BE LOVELY IF YOU COULD EXPLAIN EACH OF THESE ARCHITECTURES, E.G. WHAT DOES THE SEQUENCE MODEL DO, WHAT DOES THE LTREE DO, WHAT DOES CFG DO ETC. HOW THEY ARE DIFFERENT. 

\section{Data Collection}
To evaluate the effectiveness of the proposed methodology, two distinct datasets were used both of which follow a subject-verb-object (SVO) syntax structure: unstructured dataset and structured dataset.

The  first dataset is a subset of a multimodal \emph{SVO-probes} dataset released by Google  \cite{Google-SVO}, containing 36,841 entries. Each entry of this dataset pairs one sentence with two images—one positive and one negative. The two images have similar subjects, objects, or prepositional phrases but differ in their verbs. Only one of the images is the correct image of the sentence. An example entry is the sentence `` A dog is sitting on the road" accompanied by two images:  one with a dog sitting and another with a dog running on the road. A random subset of 350 entries of this large dataset was selected due to the run time constraints of existing quantum simulators.  We refer to this dataset as the \emph{unstructured dataset}.

\begin{figure}[h!] 
    \centering
    \includegraphics[width=1\linewidth]{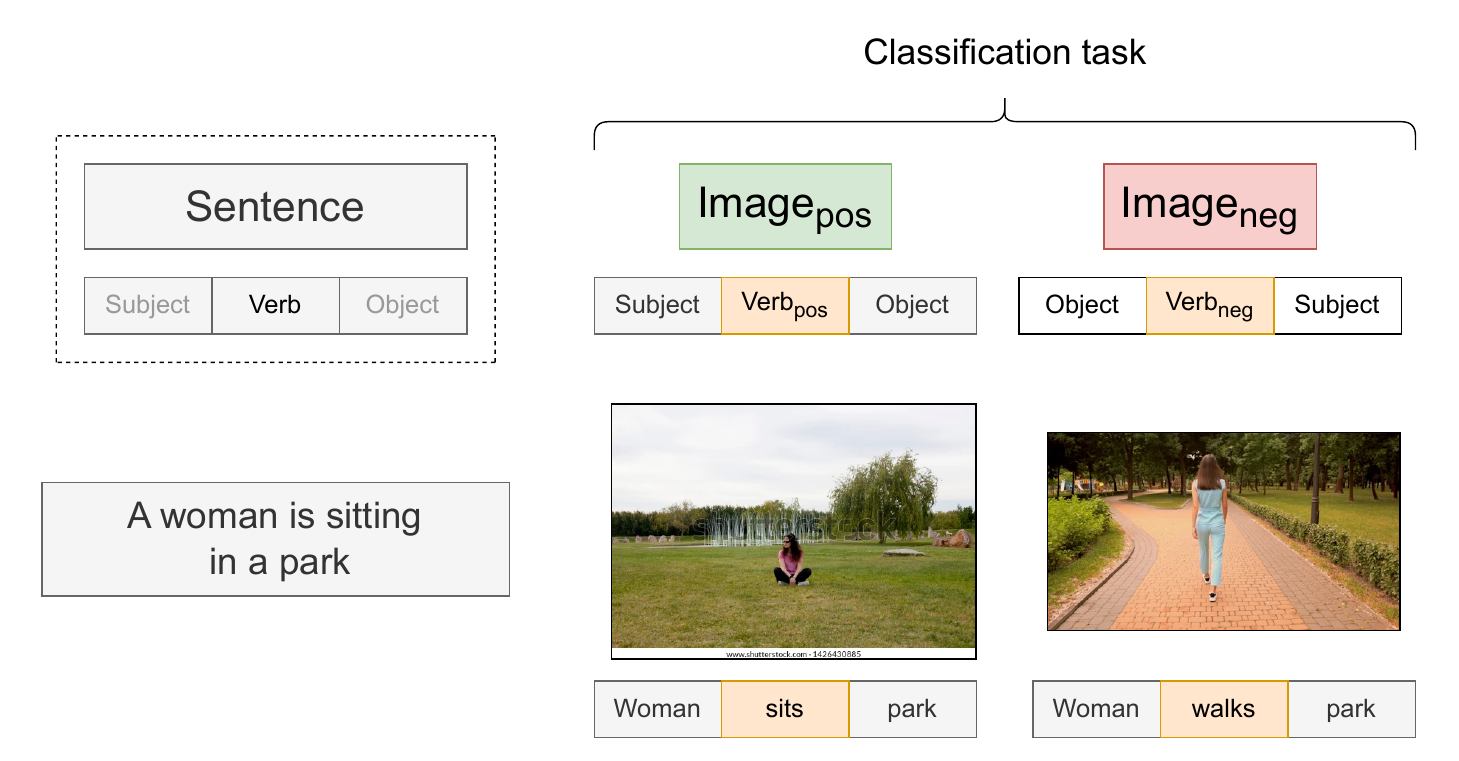}
    \caption{A high-level abstract overview of the classification task for the unstructured dataset.}
    \label{drawio-1}
\end{figure}

From this  dataset, we derived another dataset, to which we refer to as the \emph{structured dataset}. This latter  contains entries pairing two sentences with  one image. The two sentences have the same verb but their subjects and objects are swapped. An example entry  is the pair of sentences ``A child holds the mother's hand" and ``A mother holds the child's hand";   the image accompanying this pair of sentences only describes one of the sentences, e.g. is an image of  a child holding a woman's hand.  This dataset is the first one of its kind in the field. The distribution consisted of 13 different verbs each consisting of 10 different sentences. The images were downloaded from Google Search and stored in an image folder within the repository. The dataset was carefully curated to ensure an even distribution of entries and repetitive words.

Although a large number of datasets are available for multi-modal applications \cite{DatasetMultimodal}, they neither offer the transparency provided by the subject-verb-object structure nor include interchangeable subjects and objects. For guidance on the dataset size, we referred to original work conducted in \cite{QNLPinPractice}, which used a dataset size of 130.

\begin{figure}[h!] 
    \centering
    \includegraphics[width=1\linewidth]{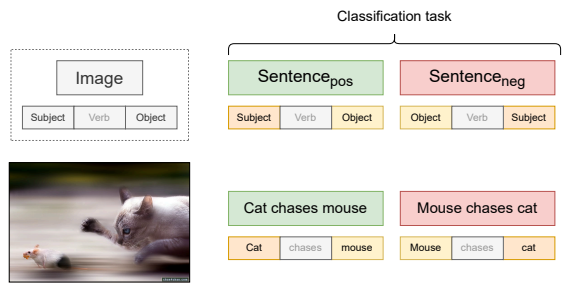}
    \caption{A high-level abstract overview of the classification task for the structured dataset.}
    \label{drawio-2}
\end{figure}

\section{Implementation}
According to the paper "\textit{QNLP in Practice: Running Compositional Models of Meaning on a Quantum Computer}", which was the first to apply the \texttt{Lambeq} toolkit, four primary approaches to training a model were identified \cite{QNLPinPractice}). For this study, we chose to implement the quantum simulation encoding both text and image data as quantum circuits, with JAX serving as the back-end to manage the training process.

Both datasets used the same image-feature extraction techniques, employing Residual Networks (ResNet) due to their ability to capture intricate details and complex patterns. We developed a \texttt{Custom16} class, inherited from PyTorch’s \texttt{nn.Module}, which customizes the ResNet-50 model. This class initializes with pre-trained weights from ImageNet and modifies the architecture by retaining all layers except the last two: the global average pooling layer and the fully connected layer. An adaptive average pooling layer downscales the feature maps to a fixed size, followed by a new fully connected layer that produces an \( n \)-dimensional feature vector suitable for quantum circuit encoding.

The \texttt{Sim14Ansatz} defines the number of qubits and layers in the quantum circuits, with 1 qubit each for \texttt{AtomicType.NOUN}, \texttt{AtomicType.SENTENCE}, and \texttt{AtomicType.PREPOSITIONAL\_PHRASE}, and 5 qubits for \texttt{image\_type}. Using 1 layer, this setup generates a 20-dimensional, non-trainable image feature vector. This isolates the impact of image features on trainable sentence parameters, enhancing flexibility in integrating image and text.

The model was built with \texttt{NumPy} and \texttt{QuantumTrainer}, using \texttt{NumpyModel.from\_diagrams} with JIT optimization for improved simulation performance.

Training utilized the Simultaneous Perturbation Stochastic Approximation (SPSA) optimizer for high-dimensional optimization, with the Binary Cross-Entropy (BCE) loss function for binary classification. Hyperparameters were set as follows: learning rate \( a = 0.02 \), perturbation factor \( c = 0.06 \), and adjustment parameter \( A = 0.001 \times \text{epochs} \), ensuring a balance between convergence speed and stability. Both datasets shared these hyperparameters, but the unstructured dataset (350 entries) trained for 200 epochs with a batch size of 20, while the structured dataset (130 entries) trained for 120 epochs with a batch size of 7. Accuracy was the primary performance metric evaluated at each epoch.

\begin{figure*}[h]
    \centering
    \begin{minipage}{0.45\textwidth}
        \centering
        \includegraphics[width=\linewidth]{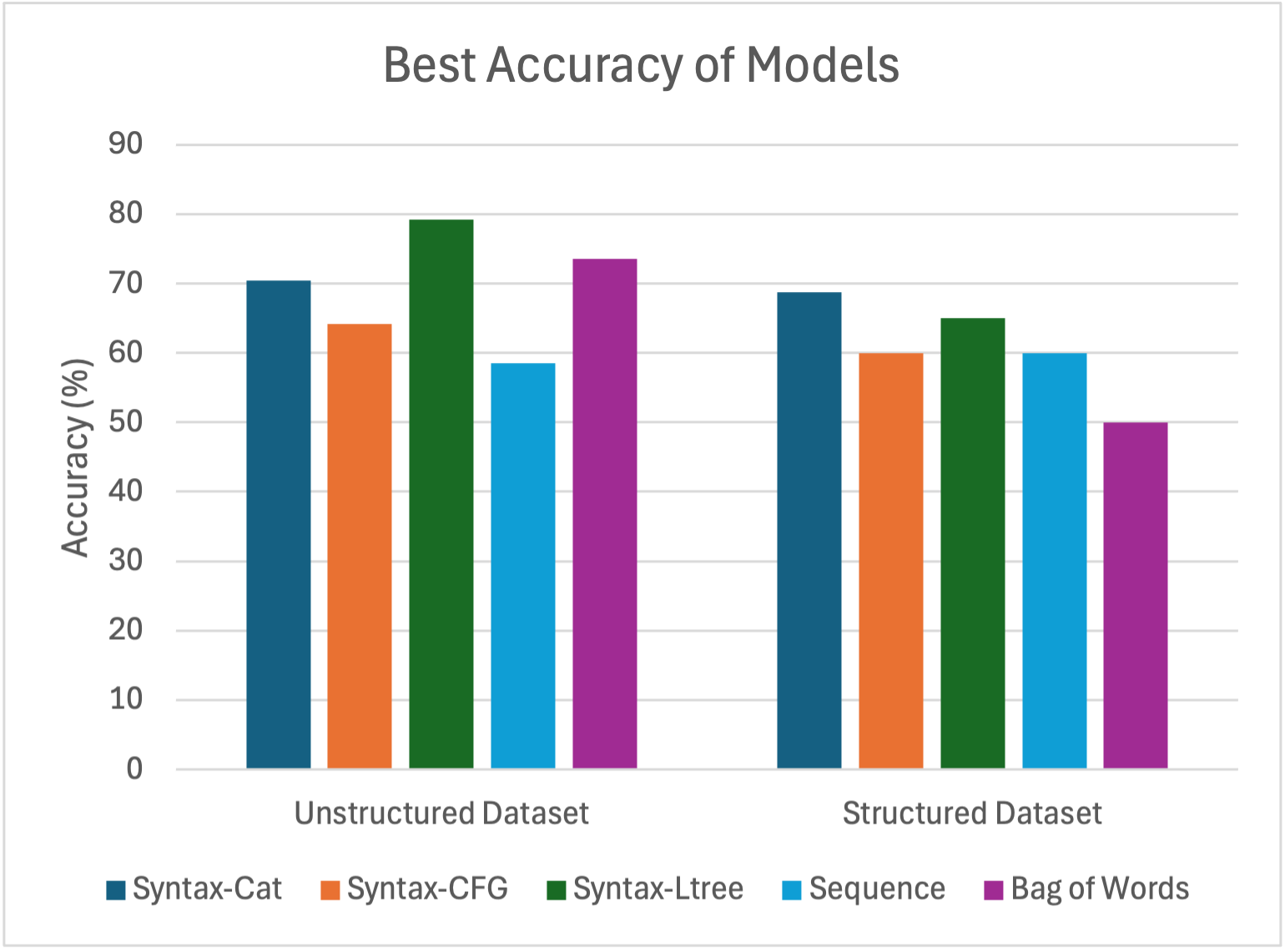}
        \caption{Bar chart representation for the best accuracy of each model.}
        \label{fig:enter-label}
    \end{minipage}\hfill
    \begin{minipage}{0.5\textwidth}
        \centering
        \renewcommand{\arraystretch}{1.1} % Adjust row height for compactness
        \setlength{\tabcolsep}{5pt} % Reduce column width to fit within the space
        \scriptsize % Set font size to a smaller size
        \begin{tabular}{|c|>{\columncolor{lightgray2}}c|c|>{\columncolor{lightgray2}}c|c|}
            \hline
            & \multicolumn{2}{c|}{\textbf{\textit{Unstructured Dataset}}} & \multicolumn{2}{c|}{\textbf{\textit{Structured Dataset}}} \\ \hline 
            \rowcolor{lightgray2}
            & \textbf{Average} & \textbf{Best} & \textbf{Average} & \textbf{Best} \\ \hline
            \textbf{Bag of Words}    & 61.88   & 73.58  & 50.00   & 50.00   \\ \hline
            \textbf{Sequence}   & 56.60   & 58.49   & 49.00   & 60.00   \\ \hline
            \textbf{Syntax-Cat} & \textbf{63.18} & 70.45   & 55.00   & 68.75   \\ \hline
            \textbf{Syntax-LTree} & 51.70    & 79.25   & 54.00   & 65.00   \\ \hline
            \textbf{Syntax-CFG}   & 60.76   & 64.15   & \textbf{56.00}   & 60.00   \\ \hline
        \end{tabular}
        \caption{Percentage accuracy results from the test set for both unstructured and structured datasets}
        \label{results_table}
    \end{minipage}
\end{figure*}

\section{Results}

\begin{figure*}[h!]
    \centering
    \begin{minipage}{0.48\textwidth}
        \centering
        \includegraphics[width=\linewidth]{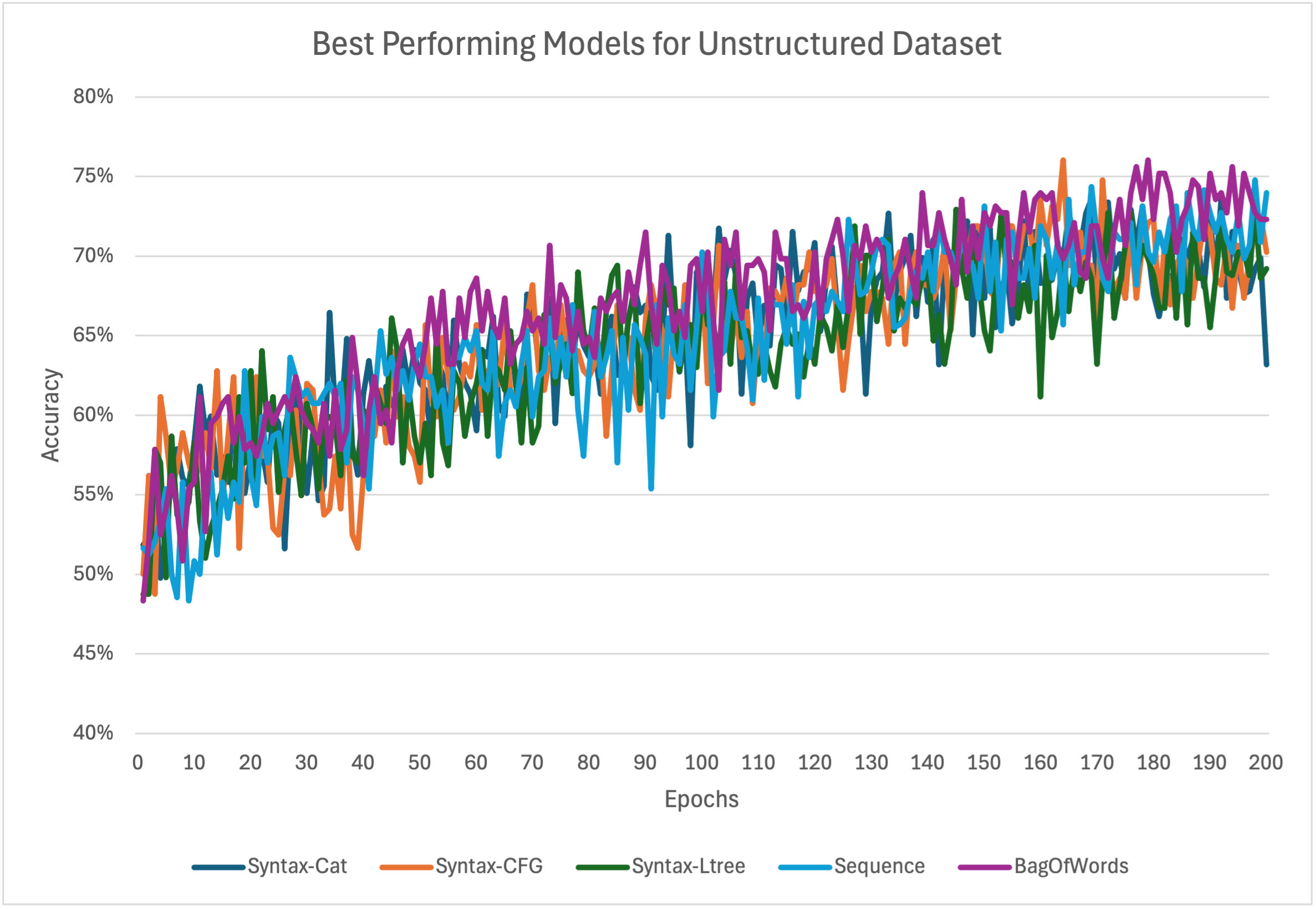}
        \caption{Line graph of the best performing models for the unstructured dataset.}
        \label{fig:graph1}
    \end{minipage}\hfill
    \begin{minipage}{0.48\textwidth}
        \centering
        \includegraphics[width=\linewidth]{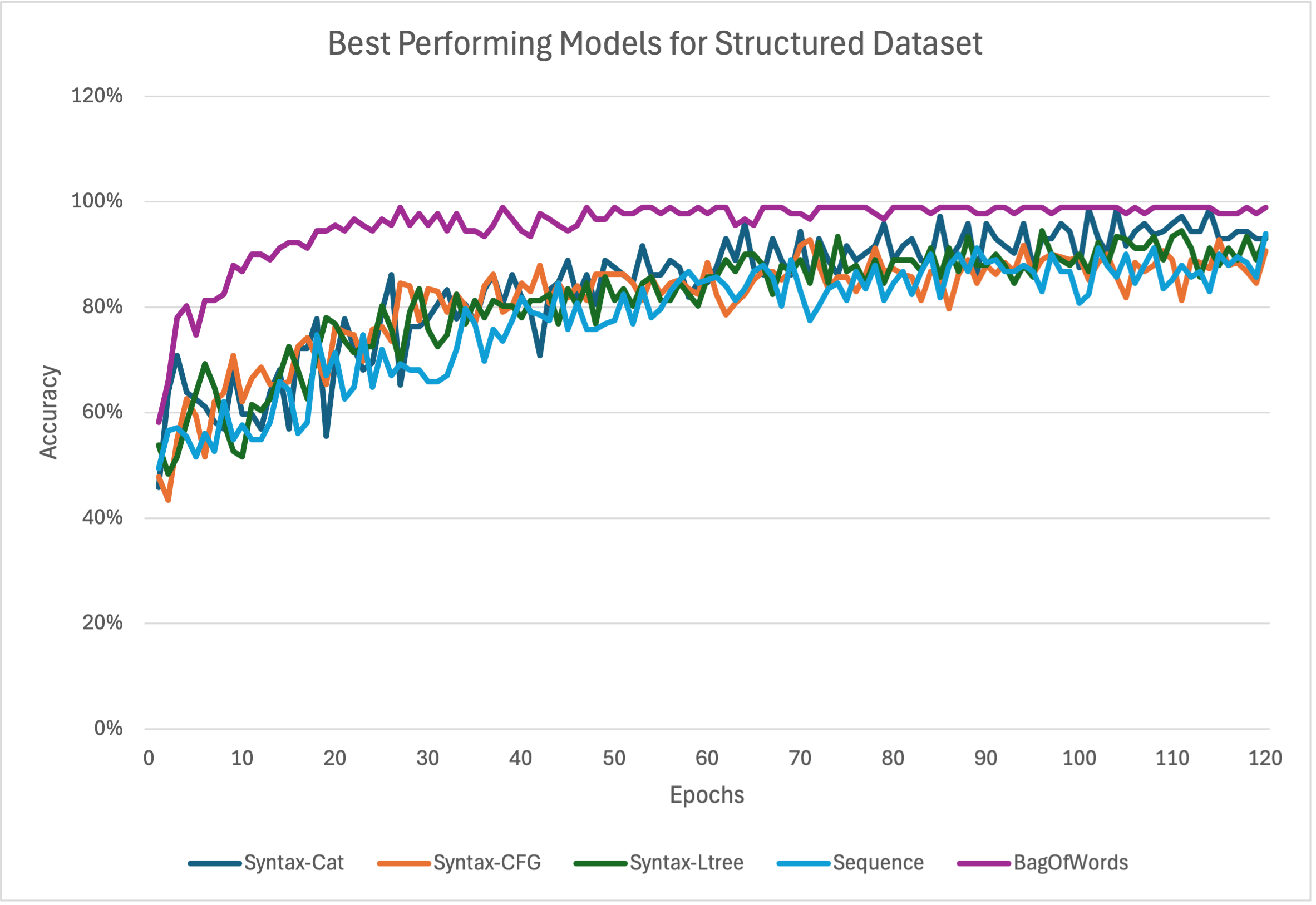}
        \caption{Line graph of the best performing models for the structured dataset.}
        \label{fig:graph2}
    \end{minipage}
\end{figure*}

Each model was trained, validated, and tested, with this process repeated five times to ensure robustness. In each iteration, the data was re-trained, re-validated, and re-tested from scratch. All models converged, some faster than others. The convergence diagrams are in Figures \ref{fig:graph1}  and \ref{fig:graph2}.

The final accuracy is reported as the average across these runs, providing a more reliable measure of predictive performance. Additionally, the highest individual accuracy achieved in any run was recorded to highlight the models' best results.

Four different compositional models were simulated, trained, and their results compared. These were three syntax-based models (see Figure \ref{fig:DisCoCat-diag}, \ref{fig:LTree-diag}, \ref{fig:CFG-diag}) and two simpler models (see Figure \ref{fig:BagOfWords-diag}, \ref{fig:WordSequence-diag}): a bag-of-words model and a word-sequence model. The syntax-based models capture the grammatical structures of sentences. In addition to our categorical model, \texttt{Cat}, we employed two other formal grammars: a recursive left-aligned tree grammar, \texttt{LTree}, and a Context-Free Grammar parse tree, \texttt{CFG}. These models aim to leverage the inherent structure in language to enhance performance. In contrast, the bag-of-words model discards word order and grammatical relations, treating sentences as unordered collections of individual words. The word-sequence model preserves word order but does not incorporate grammatical structure, offering a middle ground between simplicity and linguistic formalism. 

In the unstructured dataset, two of our three structure-aware models were top performers, with \texttt{LTree} achieving the highest individual accuracy of 79.25\%, followed by \texttt{Cat}. For average accuracy, \texttt{Cat} and \texttt{CFG} scored the highest, with the bag-of-words model ranking between them. Interestingly, the bag-of-words model performed relatively well, securing the second-best scores with only a 2\% variation. This outcome highlights its capability in basic word recognition and differentiation, especially for verbs.

Furthermore, all three structure-aware models were top performers in the structured dataset, as evidenced by both average and individual scores. This time, \texttt{Cat} achieved the highest individual accuracy at 68.75\%, followed by \texttt{LTree} and then \texttt{CFG}. In terms of average accuracy, all three syntax-based models differed by only 1\%, underscoring their comparable performance and surpassing both Sequence and Bag-of-Words by effectively managing variations in word order and syntactic relationships. Unsurprisingly, the bag-of-words model performed well in the unstructured dataset but only achieved chance-level accuracy (50\%) in the structured dataset. 

In both datasets, the bag-of-words model had the fastest convergence rate, see Figures \ref{fig:graph1}  and \ref{fig:graph2}. It converged much faster than all the other models in the structured data, and a bit faster in the unstructured one. The reason is clear since it had the simplest learning algorithm, avoiding the need to train tensors. This resulted in fewer parameters and gates, leading to faster convergence but poorer performance.

\section{Limitations}
Despite the promising results, several limitations affected the project's outcomes. One major issue was the reliance on a Numpy-based model with an SPSA optimizer, which approximates gradients by simultaneously perturbing all parameters, introducing noise and leading to the vanishing gradient problem. Although a small learning rate of 0.001 helped, the model showed limited signs of learning and did not reach stable convergence, likely impacting overall performance. A more advanced model might have mitigated these challenges, but computational constraints restricted us to simpler, less resource-intensive approaches that compromised optimal performance.

Additionally, the project's computational limits restricted the dimensionality of image vector representations. While ResNet extracted crucial features, we used 20-dimensional vectors, compared to the 2048-dimensional vectors typically employed in classical models. This significant difference reduced image representation quality, hindering the training process. Nonetheless, many models achieved over 60\% accuracy, with some reaching as high as 79\%, highlighting the potential of quantum architecture in image representation and suggesting it may capture features in ways classical models cannot.

\section{Conclusion}
In our analysis of the structured dataset, the best-performing models were the structure-aware ones, underscoring the importance of structural elements in multimodal data. As this dataset is novel, we could not compare it with existing benchmarks, but we are actively working on a large-scale variant to enhance our insights.

In the unstructured dataset, our leading model remained structure-aware, achieving an accuracy of 77.43\%. This performance exceeded that of the best classical machine learning algorithm \cite{Google-SVO} and surpassed the state-of-the-art classical categorical model, which recorded an accuracy of 78.68\% in \cite{wazniverbclip}.

While our dataset significantly surpasses the original QNLP papers, which relied on only 130 entries \cite{QNLPinPractice}, it is still relatively small compared to mainstream NLP tasks and state-of-the-art QNLP experiments, such as the 16,400-entry dataset discussed in \cite{wazni-sadrzadeh-2023-towards}. Working with larger datasets will require additional time and computational resources. To address this, we are enhancing our capabilities by utilizing the full \emph{SVO-probes} and transitioning computations to a GPU system. This upgrade will allow us to make more meaningful comparisons with the accuracies of classically trained models.

\section{Acknowledgments}

Sadrzadeh's research is supported by the Royal Academy of Engineering Research Chair/Senior Research Fellowship RCSRF2122. Both authors thank Dr. Dimitri Kartsaklis from Quantinuum Ltd. for his insights and consultations, which greatly enriched this work.

% \section{References}
\bibliography{References}  

\end{document}